\documentclass{article}

\usepackage[preprint]{corl_2026}
\usepackage[utf8]{inputenc}
\usepackage[T1]{fontenc}
\usepackage{microtype}
\usepackage{geometry}

\usepackage{amsmath, amssymb, amsfonts}

\usepackage{graphicx}
\usepackage[table,dvipsnames]{xcolor}
\usepackage{booktabs}
\usepackage{makecell}
\usepackage{nicefrac}
\usepackage{pifont}

\usepackage{enumitem}
\usepackage{caption}
\usepackage{subcaption}
\usepackage{titlesec}

\usepackage{url}
\hypersetup{
    bookmarks=true,
    colorlinks=true,
    linkcolor=black,
    citecolor=blue!50!black,
    urlcolor=blue!50!black,
}
\usepackage[capitalise]{cleveref}
\usepackage{xspace}

\usepackage{geometry}

\usepackage{tikz}

\definecolor{fetchmanblue}{HTML}{0B1849}
\newcommand{\ourname}{FetchMan\xspace}

\newif\ifshowcomments
\showcommentsfalse
\ifshowcomments
  \newcommand{\omar}[1]{\textcolor{red}{$[$#1$]^O_R$}}
  \newcommand{\maxa}[1]{\textcolor{blue}{$[$#1$]^M_A$}}
  \newcommand{\liz}[1]{\textcolor{orange}{$[$#1$]^L_Z$}}
\else
  \newcommand{\omar}[1]{}
  \newcommand{\maxa}[1]{}
  \newcommand{\liz}[1]{}
\fi

\newsavebox{\tabA}
\newsavebox{\tabB}
\newlength{\tabHt}

\title{%
  FetchMan: Learning Visual Humanoid \\
  \makebox[\textwidth][c]{Loco-Manipulation Policies from Simulated Experiences}%
}

\author{}

\begin{document}
\newgeometry{left=1.3in,right=1.3in, top=1.05in}

\maketitle
\vspace{-5em}
\begin{center}
  Omar Rayyan$^{1*}$ \quad Zhi Li$^{1}$ \quad Max Argus$^{2,3}$ \quad  Yuxin Jiang$^{1}$ \\[2mm] Chang Yu$^{1}$ \quad Chenfanfu Jiang$^{1}$ \quad Yuchen Cui$^{1}$ \\[4mm]
  $^{1}$\textbf{University of California, Los Angeles} \enspace
  $^{2}$\textbf{Allen Institute for AI} \enspace
  $^{3}$\textbf{University of Washington} \\[3mm]
  \href{https://www.orayyan.com/fetchman}{\texttt{orayyan.com/fetchman}}
\end{center}
\begingroup
\renewcommand\thefootnote{}
\footnotetext{$^{*}$Correspondence to \texttt{orayyan@ucla.edu}}
\endgroup



\begin{abstract}
Visual loco-manipulation policies that can generalize to novel scenes and objects have long been a goal of robotics research. However, today's data-hungry algorithms make collecting sufficient demonstrations a struggle for tabletop manipulation, and even more so for humanoids that must also walk and balance. Learning from simulated data and transferring that behavior to the real world, as is commonly done in locomotion, sidesteps this struggle, so we replicate that recipe for loco-manipulation. In doing so, we find that cloning synthetic demonstrations results in a low performance ceiling no matter the amount of training data. Reinforcement learning breaks through it, and refining the cloned policy with Flow-GRPO on a single sparse reward yields performance that synthetic behavior cloning cannot match. Together, these stages form our end-to-end sim-to-real pipeline spanning more than 150{,}000 scenes, which we use to train \textcolor{fetchmanblue}{\textsc{\ourname}}. We evaluate it on \textcolor{fetchmanblue}{\textsc{FetchMan-Bench}}, a simulation benchmark we release, and deploy it zero-shot on a real Unitree G1, where our single-object reach-and-pick policy walks to and grasps a target across unseen scenes at 73.3\% success. Finally, we extend this recipe to multi-object training, a first step toward loco-manipulation generalist policies at this data scale.
\end{abstract}

\vspace{4mm}

\setlength{\intextsep}{0pt}
\setlength{\textfloatsep}{0pt}
\begin{figure}[h]
    \centering
    \captionsetup{width=1.0\linewidth}
    \makebox[\linewidth][c]{%
      \includegraphics[width=1.0\linewidth]{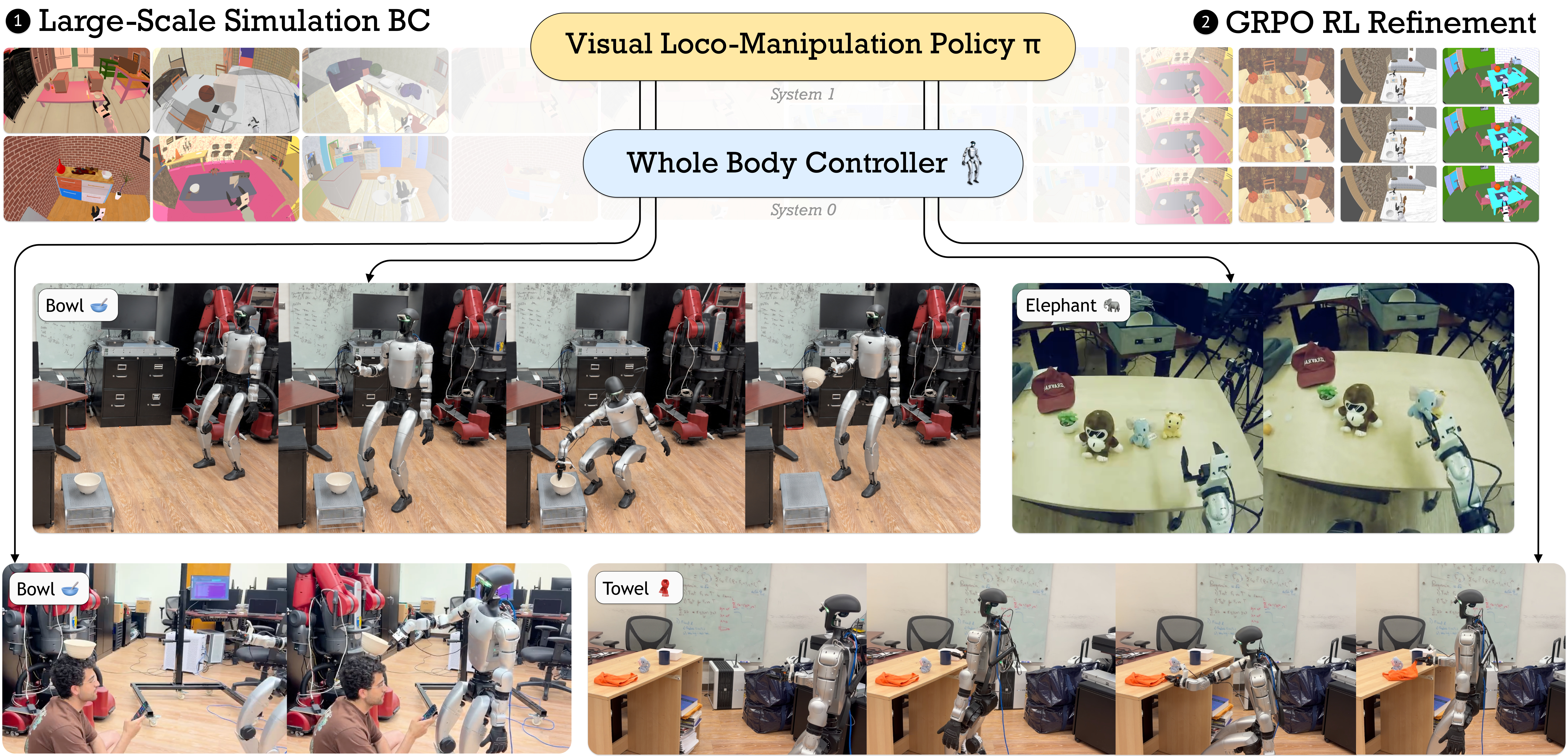}}
    \caption{\textsc{\ourname} learns visual humanoid loco-manipulation entirely in
    simulation and deploys it zero-shot to a real Unitree G1. The resulting policy
    walks up to and grasps objects across diverse, unseen real-world scenes, with no
    real-world data or per-scene tuning.} 
    \label{fig:teaser}
\end{figure}
\setlength{\intextsep}{12pt plus 2pt minus 2pt}
\setlength{\textfloatsep}{20pt plus 2pt minus 4pt}

\section{Introduction}

Simulation has become the standard setting for training humanoid locomotion policies, where interaction with the environment is typically limited to feet-floor contact and observations are low-dimensional. With easy-to-use tools~\citep{todorov2012mujoco, zakka2026mjlab, mittal2025isaac}, a walking, dancing, or back-flipping policy can be brought to life. This success has yet to translate to loco-manipulation, for two reasons. Diverse simulation assets are scarce and take significant engineering effort to scale, and contact-rich interaction is harder to model than the relatively self-contained dynamics of locomotion.

Collecting the data on real hardware instead is no escape. Scaling real-world data is hard enough for a fixed-base arm on a tabletop, and a humanoid raises both the price per trajectory and the number of trajectories required for sufficient coverage, as each demands extra supervision to ensure safe recovery and full-body balance restoration after failures. Simulation avoids these costs and further allows exploration of intermediate learning states that would be unsafe or impractical on hardware.

Recent work has begun to chip away at these limitations. MolmoSpaces~\citep{kim2026molmospaces} provides large-scale procedurally generated environments, reducing the diversity bottleneck, and MolmoBot~\citep{deshpande2026molmob0t} shows that policies trained on synthetic data from such environments transfer zero-shot to real-world tabletop manipulation under broad domain randomization. Other work trains humanoid systems purely in simulation~\citep{he2025viral, xue2025openingsimtorealdoorhumanoid}, showing that vision-based whole-body behaviors can also be deployed zero-shot to real hardware. These humanoid systems, however, are trained and evaluated within a heavily constrained visual environment structure, and rely on complex hand-crafted rewards. A vision-based humanoid loco-manipulation policy that generalizes across scene layouts without per-environment tuning remains an open problem.

Despite these advances, behavior cloning on synthetic data in simulation can be limited by two additional factors. The first is demonstration quality. Many demonstrations are inherently imperfect, and scripted simulated data in particular is prone to the imitation gap~\citep{imitation_gap, Nguyen2026CVPR}, in which the demonstrator's access to privileged information results in imperfect imitation. For loco-manipulation tasks in particular, this is caused by hidden phase variables producing discontinuous motion near the boundaries between navigation, reaching, and manipulation. And while such discontinuities could in principle be reduced, doing so demands per-task tuning that reintroduces the hand-engineering effort simulation is meant to spare us. The second factor is learning fidelity. Even given perfect demonstrations from a fully observable expert, behavior cloning cannot match that expert, because covariate shift compounds small errors into out-of-distribution states and because the imitation objective is not the task success objective.

In this work, we (1) study what scalable recipe is needed to train visual humanoid loco-manipulation policies in simulation, focusing on the factorization of locomotion and manipulation through a fixed low-level whole-body controller; (2) identify that scripted demonstration generation imposes a discrete phase structure whose boundaries are not observable from the policy's inputs and are therefore hostile to imitation, and close this gap with group-relative policy optimization (GRPO)~\citep{shao2024deepseekmath}, which improves the policy beyond the demonstrator and is enabled by simulation's identical-state group rollouts; (3) release \textcolor{fetchmanblue}{FetchMan-Bench}, a reproducible simulation benchmark for humanoid loco-manipulation; and (4) scale the resulting recipe to train \textcolor{fetchmanblue}{\textsc{\ourname}}, an environment-generalist visual loco-manipulation policy learned entirely in simulation and deployed zero-shot on real hardware, further extending it to multi-object training as an initial step toward a task generalist.

\section{Related Work}

\subsection{Synthetic Demonstration Generation}
\label{sec:rw_data}

To overcome the data scarcity that robotics faces relative to the internet-scale corpora behind language models, synthetic data has been used across a wide variety of robotics tasks, including driving~\citep{dosovitskiy17a} and navigation~\citep{spoc2023, eftekhar2024one}. Generating manipulation data, however, is more demanding, since contact-rich behaviors require high-precision physics that is harder to simulate faithfully. Harder still is pairing that fidelity with the scene diversity that visual generalization needs.

Early work~\citep{james2017transferring} showed this transfer is possible, but with limited asset and scene diversity. InternVLA~\citep{internvla_a1}, GraspVLA~\citep{GraspVLA}, and MolmoBot~\citep{deshpande2026molmob0t} later scaled up the assets, transferring zero-shot from simulation to real-world tabletop manipulation. Behavior1K~\citep{BEHAVIOR} and RoboCasa365~\citep{robocasa365} target mobile manipulation, but on a fixed set of environments rather than scaling scene diversity toward real-world transfer. We generate demonstrations at a similar physical fidelity while scaling both scene and object diversity for a humanoid embodiment, and through a mixture of imitation and reinforcement learning.

\subsection{Learning Humanoid Locomotion and Manipulation}
\label{sec:rw_humanoid}

A number of works train humanoid policies directly from teleoperation or human motion data. HumanPlus~\citep{fu2024humanplus} retargets human motion to a humanoid embodiment and combines imitation learning~(IL) with reinforcement learning~(RL) fine-tuning. OmniH2O~\citep{he2024omnih2o} enables universal whole-body teleoperation and transfers the resulting policies to real hardware. GR00T-N1~\citep{gr00tn1_2025}, Humanoid-VLA~\citep{ding2025humanoidvla}, and $\psi_{0}$~\citep{wei2026psi_0} train generalist VLA policies on heterogeneous mixtures of robot demonstrations and human video. These approaches are bottlenecked by the cost and scale of real-world data collection, which limits the diversity of scenes, objects, and failure modes they can cover, and often requires post-training adaptation at the deployment scene.

Simulation, on the other hand, enables large-scale data generation without hardware constraints. Recent systems such as VIRAL~\citep{he2025viral} and DoorMan~\citep{xue2025openingsimtorealdoorhumanoid} demonstrate that vision-based behaviors, a walk-place-grasp-turn cycle and articulated-door opening respectively, can be deployed zero-shot to real hardware. However, these policies are trained on a narrow environment structure, which limits their generalization to diverse real-world settings. ASAP~\citep{he2025asap} trains whole-body skills in simulation and refines them with real-world RL to close the sim-to-real gap, but targets a fixed set of motion skills rather than scene-level generalization.

\subsection{Post-Training Flow-Based Policies}
\label{sec:rw_rl}

Policies initialized by imitation are commonly refined with RL to surpass the demonstrator. With the increased adoption of flow-based action heads, this has become harder, as such policies have no tractable action likelihood. Some methods leave the flow untouched and instead learn a policy over its starting noise~\citep{wagenmaker2025steering}. Others like ReinFlow~\citep{zhang2025reinflow} and Flow-GRPO~\citep{liu2026flow} add Gaussian noise to the sampler, turning the deterministic ODE into a stochastic process with closed-form per-step likelihoods. These methods have largely been applied to tabletop or simulated benchmark tasks. We adopt Flow-GRPO for humanoid loco-manipulation, refining entirely in simulation across a diverse distribution of scenes to close the imitation gap left by scripted demonstrations, without any real-world interaction.

\section{Problem Setup}
\label{sec:problem}
We use the Unitree G1 humanoid with Dex1-1 grippers, exposing a single control
interface that accepts 15-dimensional whole-body commands at 10\,Hz. Only the
source of that command changes, from the scripted controller $\pi_\text{ctrl}$
(\cref{sec:data_demonstrator}) during data generation to the learned policy
$\pi_\theta$ (\cref{sec:bc}) at deployment.

\begin{figure}[t]
    \centering
    \includegraphics[width=0.99\linewidth]{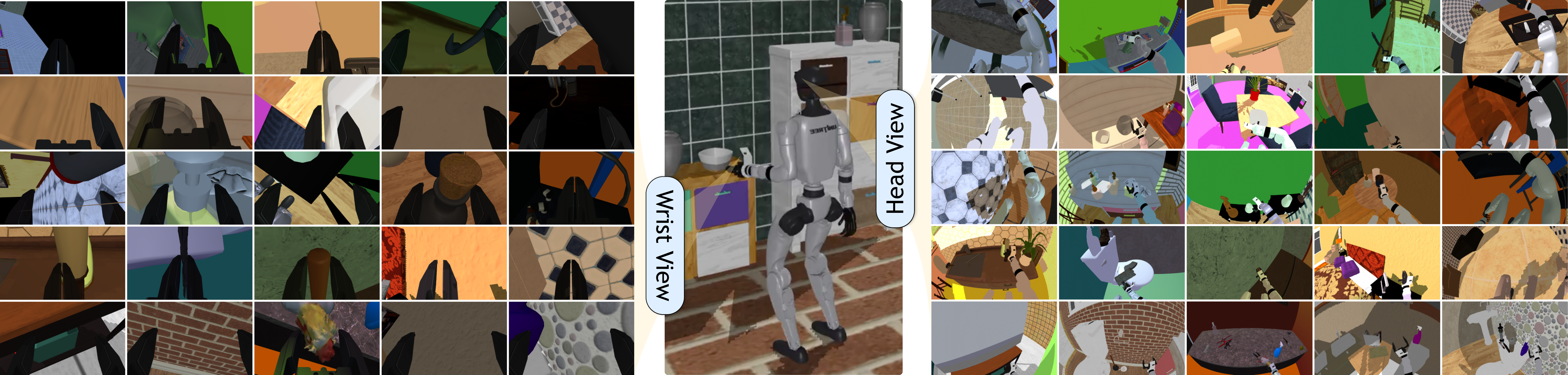}
\caption{\textbf{Simulated observations.} Example frames of the head-fisheye and
wrist streams the policy receives, across scenes randomized in texture,
lighting, and camera intrinsics and extrinsics.}
\label{fig:sim_obs}
\end{figure}

\textbf{Observations.} Our environment observation is $o_t = (I^\text{head}_t,
I^\text{wrist}_t, o^\text{prop}_t)$, where $I^\text{head}_t, I^\text{wrist}_t
\in \mathbb{R}^{224 \times 384 \times 3}$ are a head-mounted fisheye RGB image
and a wrist-mounted RGB image (\cref{fig:sim_obs}), and the proprioceptive vector $o^\text{prop}_t
= (h_t, r_t, p_t, q^\text{upper}_t, g_t)$ contains the base height $h_t$, base
roll $r_t$ and pitch $p_t$, upper-body joint positions $q^\text{upper}_t$
(waist and right arm), and the gripper aperture $g_t$.

\textbf{Actions.} The 15-dimensional action is $a_t = (v_t, \bar h_t, \bar
q^\text{waist}_t, \bar q^\text{arm}_t, \bar g_t)$, where $v_t = (v_x, v_y,
\omega) \in \mathbb{R}^3$ is a base linear and yaw velocity command, $\bar h_t
\in \mathbb{R}$ is the base-height target, $\bar q^\text{waist}_t \in
\mathbb{R}^{3}$ and $\bar q^\text{arm}_t \in \mathbb{R}^{7}$ are waist and
right-arm joint targets, and $\bar g_t \in \mathbb{R}$ is the right-gripper
target.

\textbf{Hierarchical Control.} We adopt the action factorization standard in other humanoid loco-manipulation systems~\citep{he2025viral,
xue2025openingsimtorealdoorhumanoid, wei2026psi_0}, where commands specify base
velocity, base height, upper-body joint targets, and gripper aperture. The base
command $(v_t, \bar h_t)$ is tracked by a pre-trained SONIC lower-body
controller~\citep{luo2025sonic} running at 50\,Hz, while the upper-body and
gripper targets $(\bar q^\text{waist}_t, \bar q^\text{arm}_t, \bar g_t)$ are
PD-tracked directly. Both resolve to torques at the 200\,Hz physics rate. SONIC
itself is a decoupled balance-and-walking pair, dispatched per step by a hard
threshold on the commanded base velocity. This shrinks the command space from 29
joint torques to 15 meaningful commands, which is what makes both behavior
cloning and RL fine-tuning feasible at our scale. And because the same stack
runs in simulation and on hardware, its closed-loop response is a fixed property
of the system rather than another source of sim-to-real shift.

\section{\ourname~--- Data Generation}
\label{sec:method}
We generate training data by rolling out a scripted, privileged whole-body
controller in procedurally generated indoor scenes. Each episode samples a
house, a pickable target object within it, a contact-free humanoid start pose,
and a feasible grasp, after which the controller solves the task from privileged
simulator state. The resulting trajectories pair the observations of
\cref{sec:problem} with 15-dimensional whole-body commands, and are directly
usable for imitation (\cref{fig:sample}). Below we describe how tasks are
sampled (\cref{sec:data_tasks}), the scripted demonstrator
(\cref{sec:data_demonstrator}), the mixture of episode types we collect
(\cref{sec:data_mixture}), and the per-episode domain randomization that
diversifies the data (\cref{sec:data_dr}).

\subsection{Task Sampling}
\label{sec:data_tasks}
\begin{figure*}[t]
    \centering
    \includegraphics[width=0.99\linewidth]{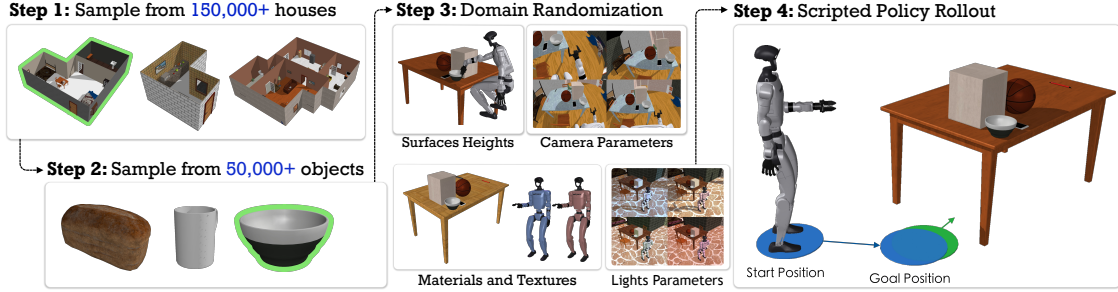}
    \caption{\textbf{Synthetic data generation pipeline.} Each episode samples
    (1)~a house from MolmoSpaces, (2)~a pickable target object within it,
    (3)~a contact-free start and grasp-standoff goal pose connected by a planned
    path the scripted controller follows, and (4)~domain randomization applied
    independently per episode.}
    \label{fig:sample}
\end{figure*}

Episodes are instantiated using MolmoSpaces~\citep{kim2026molmospaces}, an open
ecosystem of procedurally generated indoor scenes populated with physically
simulated objects. Our task sampler (\cref{fig:sample}) closely follows that
of MolmoBot~\citep{deshpande2026molmob0t}, which we extend to the humanoid
setting. A task is a triple $(s, m, x_0)$ of a scene, a target object, and a
robot start pose.

\textbf{Scene.} A scene $s$ is drawn from a pool spanning the ProcTHOR-10k,
Holodeck, and ProcTHOR-Objaverse splits curated by MolmoSpaces. Once loaded,
we precompute a 2D free-space occupancy grid $\mathcal{O}_s$ that dilates the
static collision geometry by the G1's footprint radius. $\mathcal{O}_s$ is used
both to reject infeasible base placements and to plan navigation
waypoints in \cref{sec:data_demonstrator}.

\textbf{Target object.} The target $m$ is drawn from the free-jointed assets
marked by MolmoSpaces as \emph{pickable}. Each instance carries a grasp set
$\mathcal{G}_m$, precomputed in MolmoSpaces by antipodal sampling and
post-filtering and expressed in the object frame, which we further filter by
our Dex1-1 gripper's geometry and by G1 IK feasibility. The object's pose
$T_m$ and the height of its supporting surface are resampled every episode
(\cref{sec:data_dr}).

\textbf{Robot start pose.} Candidate base positions are sampled from the free
cells of $\mathcal{O}_s$ in an annulus $r_{\min} \le \|p - p_m\|_2 \le
r_{\max}$ around the target, with yaw set to the bearing-to-target plus an
offset $\Delta \psi \sim \mathcal{U}(-0.5, 0.5)\,\text{rad}$. We accept the
first candidate $x_0$ whose whole-body pose is contact-free and from which the
target is visible in the head fisheye camera. Crucially, we validate $x_0$
against the full simulator rather than against $\mathcal{O}_s$. A flat
footprint conservatively rejects any base pose whose 2D area overlaps a
support surface, but the G1's upper body can freely occupy the volume above
that surface, e.g.,~the hand reaching over a table. Validating in 3D lets
$r_{\min}$ shrink well below what static tabletop setups assume is safe,
widening the distribution of approach geometries and better matching the
clutter of real-world deployment.

\subsection{Scripted Demonstrator}
\label{sec:data_demonstrator}
Each accepted task is rolled out by a scripted whole-body controller
$\pi_\text{ctrl}$, which acts on privileged simulator state including the object pose
$T_m$, a selected grasp $T_g$, the occupancy grid $\mathcal{O}_s$, the robot's
base pose $x_t$, and its own phase index $k_t$. None of these is visible to
$\pi_\theta$ (\cref{sec:bc}), which sees only $o_t$. The demonstrations
therefore share the action interface of \cref{sec:problem} but not its
information interface.

\textbf{Reaching.} While the robot is outside the grasp standoff,
$\pi_\text{ctrl}$ plans an A* path on an extra-inflated copy of
$\mathcal{O}_s$, reduces it to a short list of straight-line waypoints, and
emits base velocity commands $(v_x, v_y, \omega)$ that drive the robot toward
the next waypoint while keeping the chassis facing the target.

\textbf{Manipulation.} On arrival at the grasp standoff, $\pi_\text{ctrl}$
advances through a discrete phase sequence $k_t \in \{\text{reach},
\text{descend}, \text{close}, \text{lift}\}$, moving to the next once the
end-effector reaches the current phase's target. The grasp $T_g$ is selected
once per episode. We transform each candidate in $\mathcal{G}_m$ into world
coordinates through $T_m$ and take a random grasp that is reachable under
Mink~\citep{zakka10mink} inverse kinematics with self-collision avoidance
enabled. The IK treats the base height $\bar h$ as a free variable jointly optimized
with the arm joints, so $\pi_\text{ctrl}$ can drop or raise the torso to bring
$T_g$ into the arm's dexterous workspace rather than committing to a fixed
standing height. Finally, we discard episodes with invalid paths, infeasible grasps,
or phase timeouts.

\subsection{Episode Mixture}
\label{sec:data_mixture}
We collect two episode types under a fixed mixture. 80\% are \emph{fetch}
(reach+pick) episodes that exercise the full navigation-to-grasp pipeline, and
20\% are \emph{pick-only} episodes in which the robot is initialized directly at a
grasp standoff and only the manipulation phases are executed. Beyond
re-balancing the data toward the terminal grasp, the pick-only episodes let us
randomize the initial upper-body configuration, which the fetch episodes cannot.
We sample the arm joints around their defaults and draw
the initial gripper aperture uniformly between fully open and fully closed. A
fetch episode, by contrast, must arrive at the standoff from the default
arms-forward, gripper-open state, so a policy trained on those alone would
meet any deviation from it out of distribution.

\subsection{Domain Randomization}
\label{sec:data_dr}
On top of the task instance, every episode independently resamples the
following axes (\cref{fig:sample}).
\begin{itemize}[leftmargin=*,itemsep=2pt,topsep=2pt]
    \item \textbf{Textures.} Scene surfaces (walls, floors, counters, tables,
    and doors) draw textures from predefined material pools.
    \item \textbf{Lighting.} Each light source's position, color, intensity,
    and shadow toggle are resampled, and the robot's body color is tinted around
    its stock white.
    \item \textbf{Cameras.} The head fisheye and wrist cameras are perturbed
    in both intrinsics (field of view, fisheye distortion coefficients) and
    extrinsics (position and orientation offsets).
    \item \textbf{Surface and object pose.} The target's supporting surface is
    reset to a height within the G1's reachable workspace and propagated
    rigidly to every object resting on it via a contact-graph traversal. The
    target's planar pose on that surface is then jittered by a small Gaussian.
    \item \textbf{Action noise.} Small zero-mean Gaussian noise is added to
    each command before it reaches the low-level controllers, so the policy
    cannot rely on exact open-loop replay of the demonstrator.
\end{itemize}

\begin{figure}[t!]
    \centering
    \includegraphics[width=1.0\linewidth]{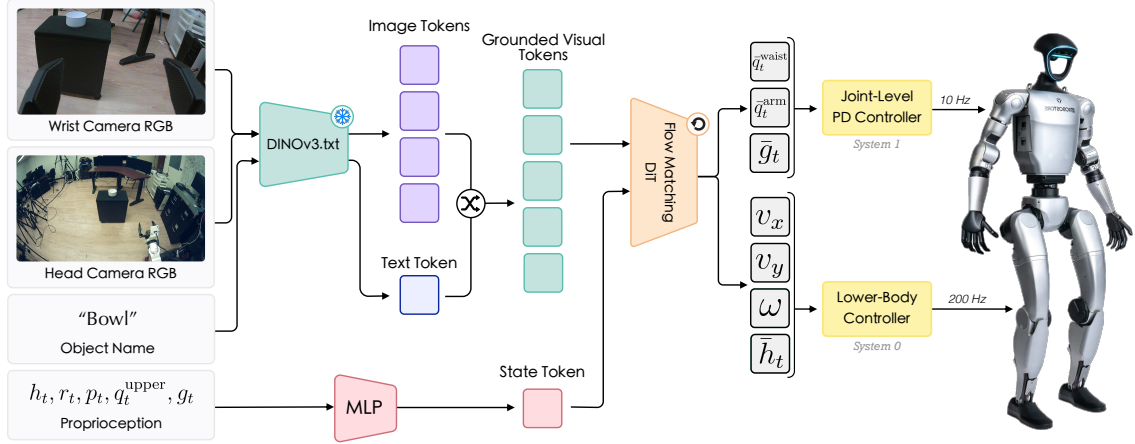}
    \caption{\textbf{Policy architecture.} The head and wrist RGB streams are encoded by a frozen DINOv3 backbone into per-patch image tokens, while the proprioceptive vector $(h_t, r_t, p_t, q^\text{upper}_t, g_t)$ is mapped by an MLP to a single state token. A flow-matching DiT cross-attends to these tokens and denoises a 15-dimensional action chunk, split into upper-body targets (waist, arm, and gripper) tracked by a joint-level PD controller and a base command $(v_x, v_y, \omega, \bar{h}_t)$ executed by the pre-trained lower-body locomotion controller.}
    \label{fig:arch}
\end{figure}

\section{\ourname~--- Policy Learning}
\label{sec:training}
\begin{figure*}[t!]
    \centering
    \includegraphics[width=1.0\linewidth]{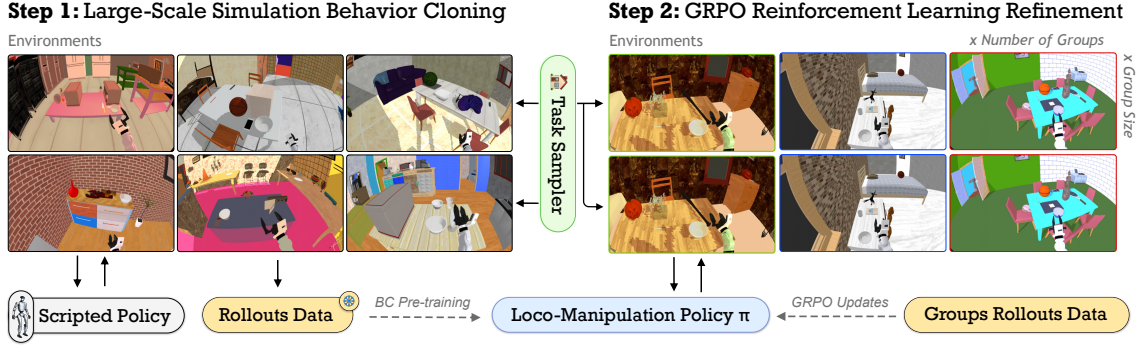}
    \caption{\textbf{Two-stage training pipeline.} \emph{Stage~1 (Imitation Pretraining).} The task sampler instantiates diverse simulated environments in which the scripted G1 staged policy generates demonstration trajectories, and the resulting frozen dataset trains the loco-manipulation policy $\pi$ by behavior cloning. \emph{Stage~2 (Online GRPO Refinement).} The same sampler resets groups of environments to identical tasks (group size $\times$ number of groups), $\pi$ collects on-policy trajectories that differ only in sampled action noise, and a group-relative GRPO update refines $\pi$ in place.}
    \label{fig:train}
\end{figure*}

We train our unprivileged policy $\pi_\theta(a\,|\,o)$ in two stages
(\cref{fig:train}). In \textbf{Stage~1}~(\cref{sec:bc}), we behavior-clone the
scripted controller $\pi_\text{ctrl}$ of \cref{sec:data_demonstrator}, giving
$\pi_\theta$ a usable prior for a high-dimensional, partially observable
control problem. In \textbf{Stage~2}~(\cref{sec:grpo}), we fine-tune
$\pi_\theta$ with large-scale on-policy RL under a sparse task-success reward.
This repairs the structural artifact that $\pi_\text{ctrl}$'s discrete phase
machine leaves on $\pi_\theta$ during behavior cloning, which forces the policy
to learn an implicit phase classifier from images and proprioception in
lieu of the hidden state $\pi_\text{ctrl}$ uses internally to advance its
phase. The resulting checkpoint is deployed zero-shot on the
real robot.

\subsection{Architecture}
\label{sec:arch}
\cref{fig:arch} summarizes our model's architecture. The vision encoder is a
frozen DINOv3 ViT-B/16~\citep{simeoni2025dinov3} that emits a patch-token grid
per camera. The proprioceptive vector $o^\text{prop}_t$ is projected into a
single state token that is appended to the visual context. The action head is
a DiT-style transformer~\citep{peebles2023scalable} that attends from noised action tokens to the observation context via cross-attention, and
predicts a chunk of actions $a_{t:t+H}$.

\textbf{Text conditioning.} For the multi-object variant of the model we
train, we swap the backbone for a frozen dino.txt~\citep{jose2025dinov2}
stack, whose vision and text towers are pre-aligned to a shared space. The
object name is encoded into a text token that cross-attention fuses into the
patch tokens, yielding \emph{grounded} visual tokens that replace them in the
context.

\subsection{Imitation Pretraining}
\label{sec:bc}
We pretrain $\pi_\theta$ by behavior cloning on the trajectories of
\cref{sec:method}. Each training sample is a window consisting of the
observation $o_t$ and the demonstrator's next $H$ action commands
$a_{t:t+H}$. Following $\pi_0$~\citep{black2024pi_0}, we re-parameterize the
11 dimensions corresponding to absolute joint or pose targets (base height,
waist, right arm) as deltas relative to a per-chunk anchor read from the
observation at $t$, leaving the base velocity command and the gripper command
absolute. We use $H=16$, and at deployment only the
first 8 actions of each chunk are executed before a new prediction is
generated from the latest observation.

\subsection{Flow-GRPO Refinement}
\label{sec:grpo}
DAgger~\citep{ross2011reduction} does not apply cleanly here, as the scripted expert acts on a phase variable it reads from privileged state, so relabeling the policy's visited states still yields actions conditioned on state the policy never observes. We therefore refine $\pi_\theta$ from the imitation checkpoint with
Flow-GRPO~\citep{liu2026flow}, the extension of group-relative policy
optimization~\citep{shao2024deepseekmath} to flow-matching policies. At
inference we replace the deterministic Euler step of the flow integrator with
a Gaussian transition $x_{\tau+\Delta\tau} \sim \mathcal{N}(x_\tau + \hat
v_\theta(x_\tau,\tau,o_t)\,\Delta\tau,\;\sigma_\tau^2|\Delta\tau|)$ whose
noise schedule $\sigma_\tau = \eta\sqrt{(1-\tau)/\tau}$ matches the
rectified-flow schedule, so the marginal at $\tau{=}1$ remains the policy's
data distribution and the per-step log-likelihood is a closed-form Gaussian
density. The same schedule is used for rollouts and for gradient passes, so
the per-step importance ratio between the current and rollout-time policies is
well defined.

Each update collects $G$ groups of $M$ episodes. Every group is reset to a
single task sample and rolled out $M$ times from that same state, so episodes within
a group differ only in the SDE noise injected by $\pi_\theta$. We use $M{=}8$
throughout, with $G{=}32$ groups ($256$ episodes per update) for the
single-object policy and $G{=}64$ ($512$ episodes per update) for the
multi-object policy of \cref{sec:exp_multiobject}. For a group with returns
$\{R_i\}$, the advantage is the standardized group-relative return $A_i = (R_i
- \mu)/\sigma$, and no learned value function is used. Groups that are entirely
successful or entirely unsuccessful carry no signal and are dropped, so only
mixed groups contribute gradient. The actor is updated with a PPO-style
clipped objective~\citep{schulman2017proximal} over the per-flow-step
importance ratio, with dual-clipping to bound the loss from below on large
negative-advantage samples, and optionally a KL penalty to the frozen behavior-cloned policy
($\beta{=}0$ for the single-object policy, $\beta{=}0.01$ for the
multi-object policy). Identical-state
group rollouts are essentially free in simulation and infeasible in the real
world, which is what makes this within-group baseline viable even at fairly
small group sizes.

The return $R_i$ is a single sparse signal, $1$ for a successful fetch of the target and $0$ otherwise, keeping the objective aligned with the
deployment metric without per-step shaping to tune across our scene and object
distribution. Since successes terminate at the lift while failures run to the
time limit, we length-normalize the actor loss by $\bar{L}/L_i$ (episode
length $L_i$, batch mean $\bar{L}$) so each episode, rather than each
transition, contributes equally and long failures do not dominate the update.

\begin{figure}[t]
    \centering
    \includegraphics[width=\linewidth]{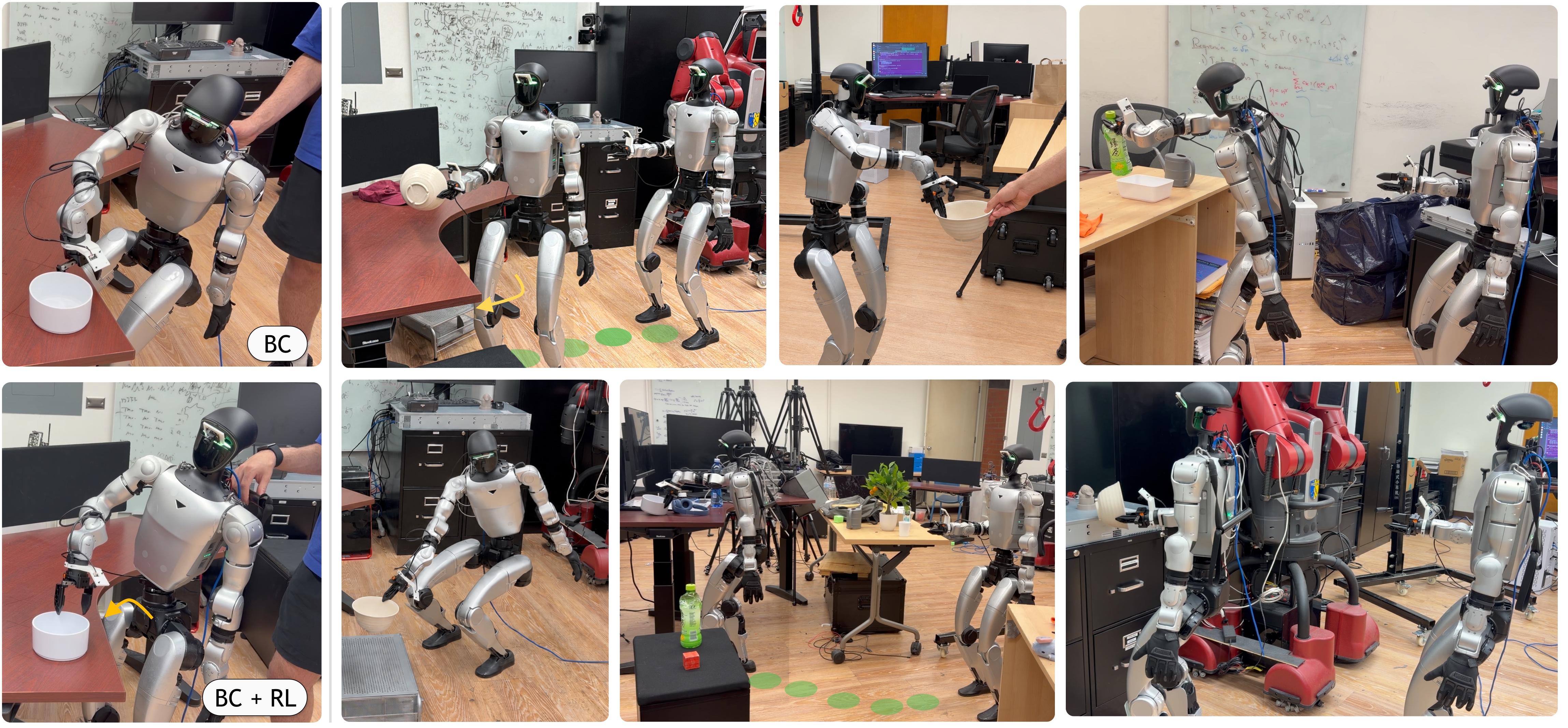}
    \caption{\textbf{Real-world rollouts and the effect of RL refinement.}
    \emph{Left.} Representative zero-shot rollouts of \ourname on the Unitree G1
    across diverse, unseen scenes, where the policy navigates to the target bowl and
    grasps it from head- and wrist-camera observations alone.
    \emph{Right.} The locomotion-to-manipulation handoff before and after
    Flow-GRPO refinement, on a fixed initialization. The behavior-cloned policy
    grasps as soon as the target is nearby and misses from too far, while after RL it
    walks in until the object is within reach and only then grasps.}
    \label{fig:rollouts}
\end{figure}

\section{Experiments}
Our experiments are organized around four questions.
\begin{enumerate}[label=\textbf{RQ\arabic*.},leftmargin=*,itemsep=2pt,topsep=2pt]
    \item Is scaling synthetic loco-manipulation demonstrations sufficient, or
    does behavior cloning saturate below the demonstrator?
    \item Does RL refinement recover the headroom that scaling synthetic data leaves behind?
    \item Does the resulting policy transfer zero-shot to a real Unitree G1,
    and which design choices are necessary for it to do so?
    \item Can the recipe extend beyond a single object toward
    object-generalist loco-manipulation?
\end{enumerate}

\begin{figure}[t]
    \centering
    \begin{minipage}{0.48\linewidth}
        \centering
        \includegraphics[width=\linewidth]{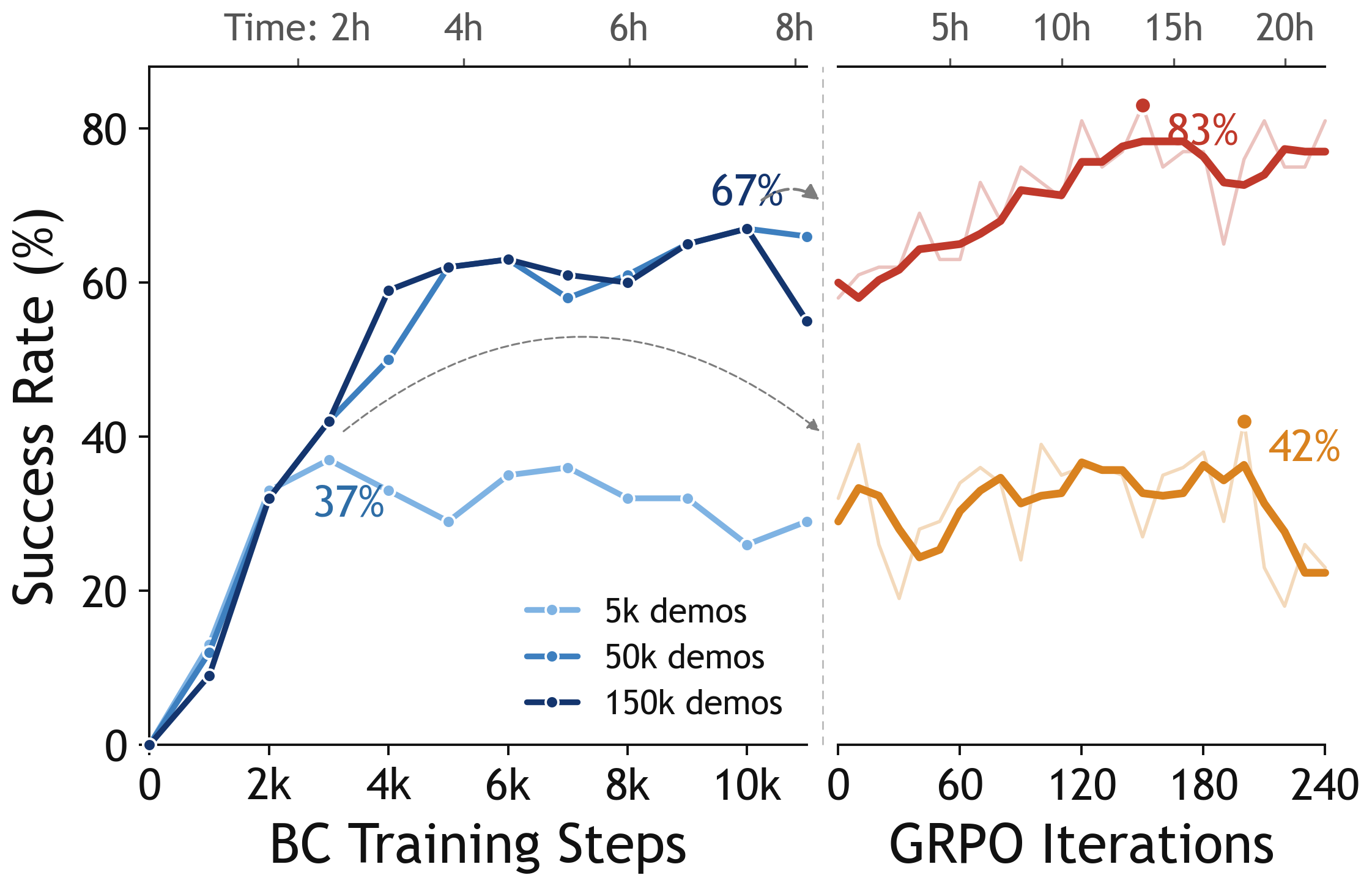}
    \end{minipage}\hfill
    \begin{minipage}{0.48\linewidth}
        \centering
        \includegraphics[width=\linewidth]{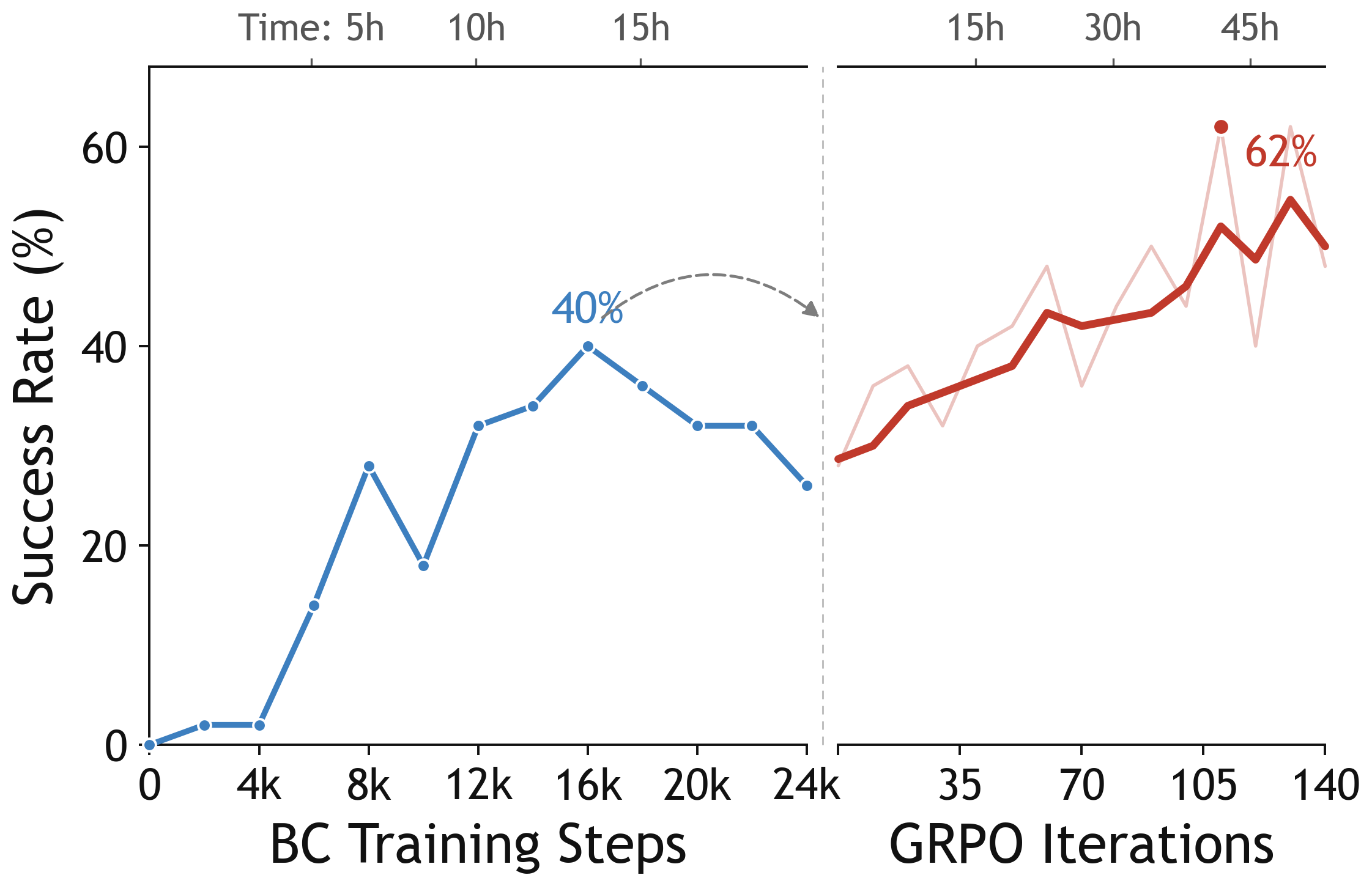}
    \end{minipage}
    \caption{\textbf{Behavior cloning saturates, RL breaks through.}
    \emph{Left.} Single-object loco-manipulation SR (\%) on the 100-scene
    validation set. The first segment scales BC demonstrations (5k--150k) and
    plateaus after 50k, while the second continues the 150k checkpoint with GRPO,
    which pushes past the imitation ceiling. A second run applies the same
    refinement to the 5k checkpoint and does not improve comparably.
    \emph{Right.} The text-conditioned multi-object policy, whose BC success
    rate peaks near 40\% before GRPO lifts it to roughly 62\%. Time in GPU hours.}
    \label{fig:scaling}
\end{figure}

\subsection{Setup}
\label{sec:exp_setup}
Instantiating the data-generation pipeline of \cref{sec:method}, we collect
150k bowl-pick demonstrations, roughly 650 hours of robot experience across
$\sim$150k distinct procedurally generated scenes, at $\sim$100 episodes per
minute on a single NVIDIA L40S ($\sim$40 GPU-hours in total). Unless stated
otherwise, our reference model uses the frozen DINOv3 encoder, delta-action
targets, and the full 150k-demonstration set. We call it \ourname-BC before
RL refinement and \ourname~(BC+RL) after.

\subsection{FetchMan-Bench}
\label{sec:benchmark}
We evaluate on \textsc{FetchMan-Bench}, our reproducible simulation benchmark for
humanoid loco-manipulation, with fixed held-out scenes and scoring released so
results are comparable across methods. It reports two success rates.
\emph{Manipulation} SR isolates the terminal phase, where the robot is initialized
at a grasp standoff and must grasp and lift the target. \emph{Loco-manipulation}
SR measures the full task, in which the robot starts away from the object and
must navigate to it before grasping, additionally stressing the
locomotion-to-manipulation handoff at which $\pi_\text{ctrl}$ crosses its
hidden phase boundary. In simulation we evaluate on a fixed set of 100
initializations never seen during training and report the best validation
checkpoint, which is then deployed zero-shot on the physical G1. No real-world data is used at any stage. Standard errors assume a binomial distribution.

\subsection{Scaling Alone Saturates Below the Demonstrator (RQ1)}
\label{sec:exp_scaling}
We first ask whether more demonstrations are enough. The left segment of
\cref{fig:scaling} scales the behavior-cloning set from 5k to 150k
demonstrations. Loco-manipulation SR climbs steeply from 5k to 50k (40\% to
67\%) and then \emph{saturates}, as the further jump to 150k yields no measurable
gain (67\%). This plateau is the empirical signature of the imitation ceiling
argued in \cref{sec:training}. Because $\pi_\text{ctrl}$'s phase boundaries are
driven by a hidden index $k_t$ that the policy cannot observe, additional
demonstrations from the same demonstrator cannot teach the policy what it
structurally cannot see. Scaling data alone is therefore insufficient.

In principle this ceiling could be raised by engineering a demonstrator with
smoother, more varied phase transitions, but doing so reintroduces the
per-task, hand-tuned controller design that scaling is meant to
avoid. Our RL stage instead requires only a single sparse success reward. We therefore
treat the ceiling as intrinsic to any fixed scripted demonstrator and move
past it with reinforcement learning (\cref{sec:exp_rl}).

\subsection{RL Refinement Recovers the Headroom (RQ2)}
\label{sec:exp_rl}
The right segment of \cref{fig:scaling} continues the saturated 150k
checkpoint with Flow-GRPO. RL breaks through the plateau, lifting simulated
loco-manipulation SR from 67\% to 83\%, confirming that the remaining headroom
comes from a different learning signal rather than from more data, one that
lets the policy relax the demonstrator's discrete phase boundaries rather than
imitate them.

The same refinement is far less effective from a weaker initialization. The
second trajectory in \cref{fig:scaling} applies identical Flow-GRPO settings
to the 5k-demonstration checkpoint (40\%), and over the same number of
iterations it oscillates around its starting success rate, peaking at 42\%
without a clear upward trend. RL refinement therefore does not substitute the scale of imitation pretraining needed. The group-relative advantage needs a
prior that already succeeds often enough for groups to be mixed, and a policy
near the floor yields mostly all-failure groups that are dropped by the
filter.

\cref{tab:main} isolates where these gains land. RL raises loco-manipulation
SR in both simulation (67\% to 83\%) and the real world (56.7\% to 73.3\%),
while manipulation SR is nearly unchanged (72.7\% to 77.2\% in the real
world). The terminal grasp is already well cloned, and RL's gains concentrate
almost entirely in the loco-manipulation regime, exactly the phase-transition
behavior that \cref{sec:training} predicts behavior cloning cannot recover
from privileged-state-driven demonstrations.

\begin{table}[t]
\captionsetup{font=scriptsize}
\setlength{\abovecaptionskip}{2pt}
\sbox{\tabA}{\resizebox{0.49\textwidth}{!}{%
\begin{tabular}{lcccc}
\toprule
& \multicolumn{2}{c}{Simulation SR [\%]} & \multicolumn{2}{c}{Real-World SR [\%]} \\
\cmidrule(lr){2-3}\cmidrule(lr){4-5}
Method & Manipulation & \shortstack{Loco-\\Manipulation} & Manipulation & \shortstack{Loco-\\Manipulation} \\
\midrule
Num. Trials & 100 & 100 & 22 & 30 \\
\midrule
\ourname-BC      & $75.0 \pm 4.3$ & $67.0 \pm 4.7$ & $72.7 \pm 9.5$ & $56.7 \pm 9.0$ \\
\ourname~(BC+RL) & $\mathbf{79.0 \pm 4.1}$ & $\mathbf{83.0 \pm 3.8}$ & $\mathbf{77.2 \pm 8.9}$ & $\mathbf{73.3 \pm 8.1}$ \\
\bottomrule
\end{tabular}}}
\sbox{\tabB}{\resizebox{0.49\textwidth}{!}{%
\begin{tabular}{lcccc}
\toprule
& \multicolumn{2}{c}{Simulation SR [\%]} & \multicolumn{2}{c}{Real-World SR [\%]} \\
\cmidrule(lr){2-3}\cmidrule(lr){4-5}
Variant & Manipulation & \shortstack{Loco-\\Manipulation} & Manipulation & \shortstack{Loco-\\Manipulation} \\
\midrule
\ourname-BC~(DINOv3, delta) & $\mathbf{75.0 \pm 4.3}$ & $\mathbf{67.0 \pm 4.7}$ & $\mathbf{72.7 \pm 9.5}$ & $\mathbf{56.7 \pm 9.0}$ \\
\quad with SigLIP            & $58.0 \pm 4.9$ & $42.0 \pm 4.9$ & $8.3 \pm 8.0$  & $0.0$ \\
\quad with absolute actions  & $62.0 \pm 4.9$ & $45.0 \pm 5.0$ & $16.7 \pm 10.8$ & $0.0$ \\
\bottomrule
\end{tabular}}}
\setlength{\tabHt}{\ht\tabA}\addtolength{\tabHt}{\dp\tabA}
\setlength{\dimen0}{\ht\tabB}\addtolength{\dimen0}{\dp\tabB}
\ifdim\dimen0>\tabHt \setlength{\tabHt}{\dimen0}\fi
\begin{minipage}[t]{0.49\linewidth}
\vspace{0pt}
\centering
\usebox{\tabA}\par
\vspace{\dimexpr\tabHt-\ht\tabA-\dp\tabA\relax}
\caption{\textbf{Main results.} RL refinement improves the full
loco-manipulation task in both simulation and the real world while leaving the
already-saturated terminal grasp essentially unchanged. Values are success
rates (SR) in [\%] $\pm$ standard error.}
\label{tab:main}
\end{minipage}\hfill
\begin{minipage}[t]{0.49\linewidth}
\vspace{0pt}
\centering
\usebox{\tabB}\par
\vspace{\dimexpr\tabHt-\ht\tabB-\dp\tabB\relax}
\caption{\textbf{Architecture ablations.} A frozen DINOv3 encoder and
delta-action targets are each necessary for zero-shot transfer; replacing
either collapses real-world performance. Values are success rates (SR) in
[\%] $\pm$ standard error, over the reference BC configuration. Simulation
is over 100 held-out initializations; real-world trials are 22 manipulation
and 30 loco-manipulation for the reference, and 12/10 respectively for
the ablation variants.}
\label{tab:ablation}
\end{minipage}
\end{table}

\subsection{Zero-Shot Transfer and What Enables It (RQ3)}
\label{sec:exp_main}
The reference checkpoint is deployed zero-shot on the physical G1, with no
real-world data at any stage. \cref{fig:rollouts} shows representative
rollouts across diverse, unseen scenes, where the policy navigates to the target and
grasps it from head- and wrist-camera observations alone. The final
\ourname~(BC+RL) model reaches 73.3\% loco-manipulation SR on hardware
(\cref{tab:main}), establishing that the simulation-trained recipe transfers.

\cref{tab:ablation} identifies what makes this transfer possible, holding
everything else at the reference configuration. Two choices dominate. The first
is the vision encoder. A frozen DINOv3 encoder substantially outperforms SigLIP
in simulation (67\% vs.\ 42\%), and in the real world SigLIP collapses almost
entirely (8.3\% manipulation, 0\% loco-manipulation). We attribute this to
DINOv3's stronger geometric features, which cross the sim-to-real appearance
gap without finetuning. The second is the action parameterization. Absolute
targets, trained for the same number of
steps, reach only 16.7\% manipulation and 0\% loco-manipulation on hardware,
confirming that the near-zero-mean delta targets helps in the transfer.

\subsection{Toward Object Generalism (RQ4)}
\label{sec:exp_multiobject}

\begin{figure*}[t!]
    \centering
    \includegraphics[width=1.0\linewidth]{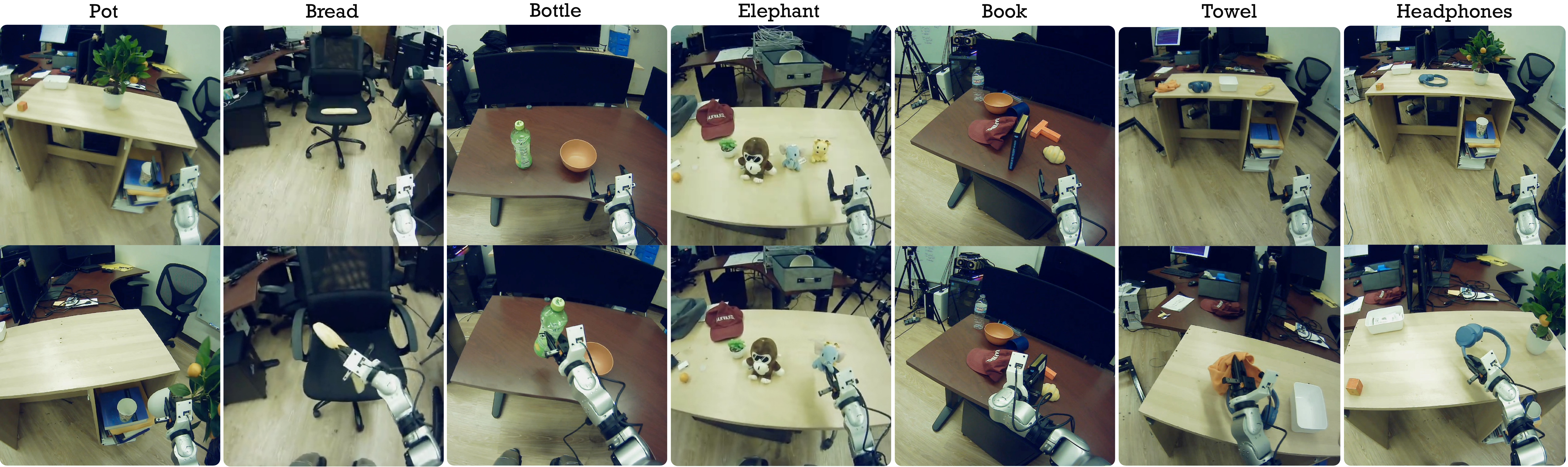}
    \caption{\textbf{Multi-object rollouts.} Head-camera views from the
    text-conditioned policy grasping different target objects, one episode per
    row, with the prompted object name shown at left.}
    \label{fig:multiroll}
\end{figure*}

Finally, we ask whether the same recipe reaches beyond a single object. We
regenerate data across all pickable object categories, collecting 350k
demonstrations under an identical pipeline, and train the text-conditioned
variant of \cref{sec:arch} that grounds a dino.txt backbone on the target
object's name. We behavior-clone and then refine this model with Flow-GRPO, as
shown in \cref{fig:scaling}, and \cref{fig:multiroll} shows example rollouts
across object categories. Because this policy's behavior-cloned prior succeeds
less often than the single-object one, a larger fraction of groups return
entirely unsuccessful and are discarded by the mixed-group filter of
\cref{sec:grpo}, so we double the group count to $G{=}64$ to retain a
comparable number of gradient-carrying groups per update. In simulation, the
multi-object BC policy reaches 40\% and the fine-tuned policy 62\%
loco-manipulation SR, below the single-object performance of 83\%. Some of this
difference is expected given the more difficult task
distribution.\footnote{On its own data, which used a similar object
distribution, MolmoBot reports 100\% success on bowls against 66.8\% on diverse
objects.}

The lower performance of the base multi-object policy relative to the BC-only
MolmoBot~\citep{deshpande2026molmob0t} policy suggests that substantial gains could come from switching from
an LBM-style~\citep{barreiros2026careful} architecture to a VLA-style one. The dino.txt backbone is frozen
and comparatively small, and a larger pre-trained vision-language model used as
the encoder would bring stronger grounding of the target name to its visual
referent. Finetuning the backbone jointly with the policy at a low learning
rate is a further option. We did not pursue these strategies, as they are
substantially more computationally expensive and our compute and hardware
access were limited.

We did, however, run qualitative zero-shot evaluations on the real G1. The
policy produces successful fetches across several object categories, though it
is noticeably less robust than the single-object policy. This answers our
research question positively, in that the recipe can extend to the multi-object
setting, with further scaling left to future work. Videos are shared on our website.

\section{Limitations}
\label{sec:limitation}
\textbf{Stateless policy.} Our $\pi_\theta$ acts on a single observation $o_t$
with no history, so the demonstrator's phase index $k_t$ must be inferred from
one frame. A short observation history would likely make the phase partially
recoverable, but stacking frames multiplies the visual token count and the cost
of every training step, and we did not have the compute budget to ablate it
here.

\textbf{Fixed lower-body controller.} $\pi_\theta$ commands SONIC through base
velocity and height and never adapts it. This decoupling makes learning
tractable at our scale but fixes the robot's gait and balance behavior. The
policy cannot brace against heavy objects, widen its stance to extend reach,
or recover from disturbances outside SONIC's envelope. Unfreezing the lower
body would return the policy to a 29-dimensional torque space and require
relearning balance alongside manipulation, at a scale cost beyond our budget.

\textbf{Task diversity.} Our tasks are limited to fetch (reach-and-pick). MolmoSpaces
supports articulated-object manipulation, and longer-horizon behaviors could be
composed from atomic skills of this kind. We leave this to future work.

\section{Conclusion}
\label{sec:conclusion}
We presented \ourname, which learns visual humanoid loco-manipulation entirely
in simulation and deploys it zero-shot to a real Unitree G1. A scripted
controller generates demonstrations across hundreds of thousands of scenes and
objects, a flow-matching policy is behavior-cloned on them, and group-relative
RL refines it beyond the demonstrator. Behavior cloning saturates well below
the achievable ceiling, which we attribute to the demonstrator's unobservable
phase structure, and RL recovers the remaining headroom in both simulation and
the real world. A preliminary text-conditioned model extends the recipe to
multiple object categories, taking a first step toward object generalism.

\section*{Acknowledgments}
\label{sec:ack}
We thank Nirvana for providing the Unitree G1 robot used in our experiments, and Mahi Shafiullah for helpful discussions throughout. This work is partially supported by DARPA SAFRON grant number HR0011-25-3-0141, and gifts from Cisco Research. Approved for public release; distribution is unlimited.

\bibliography{references}

\end{document}